\documentclass{sig-alternate}
\usepackage{breqn}
\usepackage{amsmath}
\usepackage{graphicx}

\begin{document}
%
\conferenceinfo{IBM I-CARE}{'14 October 9-11,2014, Bangalore,India}

\title{Characterizing driving behavior using automatic visual analysis}
\subtitle{
}
%
%
%
%
%

\numberofauthors{2} 
%
\author{
%
%
\alignauthor
Mrinal Haloi\titlenote{Mrinal Haloi B.tech Student}\\
       \affaddr{IIT Guwahati}\\
       \email{h.mrinal@iitg.ernet.in}
\alignauthor
Dinesh Babu Jayagopi\titlenote{Assistant Professor.}\\
       \affaddr{IIIT Bangalore}\\
       \email{jdinesh@iiitb.ac.in}
}

\maketitle
\begin{abstract}
\textbf{
In this work, we present the problem of rash driving detection algorithm using a single wide angle camera sensor,
 particularly useful in the Indian context. To our knowledge this
rash driving problem has not been addressed using Image
processing techniques (existing works use other sensors such
as accelerometer). Car Image processing literature, though
rich and mature, does not address the rash driving problem. In this work-in-progress paper, we present the need to
address this problem, our approach and our future plans to
build a rash driving detector.}
\end{abstract}

\category{H.4.3}{Information Systems Applications}[Communications Applications]

\terms{ADAS,Camera sensor,Image analysis}

\keywords{Rash Driving detection, Autonomous Driving}

\section{Introduction}
India is one of the most accident prone country, where according to the NCRB report 135,000 died in 2013 
and property damage of worth \$ 20 billion \cite{report2013}. Many a time, accidents 
and unusual traffic congestion take place due to careless and impatient nature of drivers. 
In most cases drivers don't follow lane rules, traffic rules leading to traffic congestion 
and accidents. Taking effective measure on traffic situation \cite{hong2007,baehring2005} 
and driver behaviour \cite{trivedi2005} can prevent accidents and congestion.

A developing country like India needs an effective traffic monitoring and management system. 
Towards this we propose a visual-analysis-based driving behaviour monitoring system. The 
visual analysis includes the acceleration, lane-changing, and distance-maintaining behaviour
(both from a near-by car and pedestrians). To enhance traffic safety, making road accident 
and congestion free, cab companies including government and private can adapt our system. 
Public transportation department can install this system in buses and other vehicles and 
in traffic junction for monitoring.

In developed countries like U.S.A, with the gradual emergence of autonomous driving 
research, efforts are on to build a smart driving system that can drive more safely without 
any fatigue, as compared to humans can be programmed to follow traffic rules. Even for these 
automatic cars, modelling self-driving behavior by considering distances of surrounding cars 
and detecting pedestrians is very relevant. 

In this work, we use a single wide angle camera sensor for capturing surrounding environment 
for visual analysis of other drivers behaviour and detecting nearby obstacles. From this data,
informative features namely fast side-ways and forward acceleration, wrong direction driving, 
frequent lane changing, getting-close-to-other-cars-and-pedestrians behavior are computed by using 
visual analysis techniques. From this collection of features, rash driving behavior can be detected. 
In Figure 1 we have depicted different possible scenarios for rash driving detection using camera 
on different infrastructures. 

\begin{figure}[h!]
  \centering
      \includegraphics[width = 2.7in,height=1.5in]{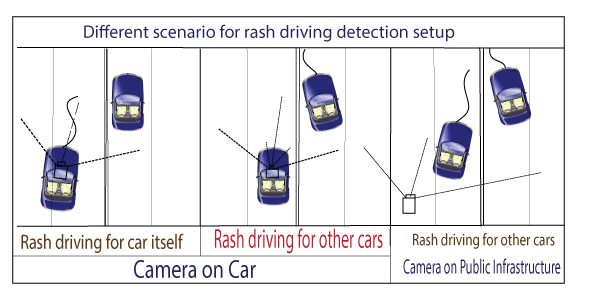}
  \caption{Rash driving Scenarios}
\end{figure}

The solution to the problem of rash-driving detection using visual analysis is a novel contribution
(as compared to \cite{dhar2014}). Also, it is a socially-relevant problem in the Indian context. So far, we 
have defined and extracted the relevant visual features on publicly available datasets. We have 
collected a small sample of data in the city of Bangalore for some intial experiments. In the 
future, we plan to record more videos by tieing-up with professional drivers to collect a new
dataset, to test and advance this initial approach. We also plan to work with government
agencies who are interested in sharing the data of traffic junctions in Bangalore.


\section{Related Work}
The related literature can be classified into three categories. 
First, the works on Image Processing and Computer Vision using single or multiple camera facing the road. 
Second, driver behaviour understanding using a camera facing the driver. 
Finally, a limited literature on rash-driving, albeit not using Image Processing. 

In the first category, we have works on advanced driver assistance system, traffic safety,
autonomous vehicle navigation and driver behaviour modelling using mutiple 
cameras, LIDAR, RADAR sensor etc. These works focus on using image processing and learning
based method for lane detection, road segmentation, traffic signs detection and recognition, 
3D modelling of road environment (e.g. \cite{dataset2013, baehring2005, hong2007, srinivasa2003})
Parallax flow computation was used by Baehring et al. for detecting overtaking and close 
cutting vehicles \cite{baehring2005}. For detecting and avoiding collison, Hong et al. 
had used Radar, LIDAR, camera  and  omnidirectional camera repectively \cite{hong2007,hong20071}. 
They focused on detecting using LIDAR sensor data classifying object 
as static and dynamic and tracking using extended Kalman filter and for getting a wide view 
of surrounding situation. For detection of forward collision Srinivasa et al. have used 
forward looking camera and radar data \cite{srinivasa2003}. 

Regarding the second and third category, the literature is fairly limited. In some works 
driver inattentiveness was modelled using fatigue detection, drowsiness, eye tracking, cell phone 
usage etc. Ji et al.\cite{ji2002} presented tracking method for eye, gaze and face pose and Hu et al.\cite{hu2009} used SVM based method for driver drosiness detection . 
Trivedi et al modelled driver behavior using head movements for detecting driver gaze and distraction, 
targetting adavanced driver safety \cite{trivedi2014,trivedie2014}. 
Using accelerometer and orientation sensor
data \cite{dhar2014}, rash driving warning system was developed as a mobile application.

\section{Characterizing rash driving}
Rash drivers generally tend to accelerate quickly side-ways and in forward direction. They change lanes frequently and get dangerously close to others vehicles and people. In this section we describe our rash driving estimation algorithm, as visualized in Fig.[2]. From video we take two consecutive frames for extracting features. This features will acts as a input for rash driving algorithm which will be based on thresholding of features values. If rash driving detected we will extract number plate of the car, otherwise will run this algorithm for next consecutive frames. 

\begin{figure}[h!]
  \centering
      \includegraphics[width=0.5\textwidth]{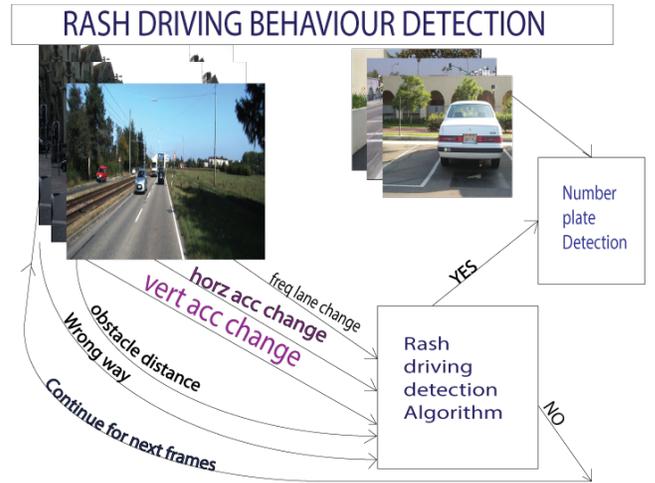}
\caption{Our rash driving detection algorithm}
\end{figure}

\subsection{Fast side-ways and forward Acceleration}
Rapid acceleration also contributes to rash driving. By computing 
optical flow we can estimate horizontal and 
vertical flow change of road environment. Frequent change in horizontal 
flow in the regions of detected cars result of rash lane changing and vertical 
flow change can give knowledge about relative velocity change of test car with 
respect to surrounding cars. From optical flow of surrounding region we  
predict the rash behaviour of other cars. The exact procedure for computing
the discrete flow is described below. 
\subsubsection{Discrete Flow Computation}
Optical flow is a measure of pixel velocity in two frame of a video. Below we have presented objective function for optical flow \cite{black2014} computation.
\begin{dmath}
E(u,v) = \displaystyle\sum\limits_{i,j} {\rho_{D}(I_{1}(i,)) - I_{2}(i + u_{i,j},j+v_{i,j})}+ \lambda[\rho_{s}(u_{i,j}-u_{i+1,j}) + \rho_{s}(u_{i,j}-u_{i,j+1}) + \rho_{s}(v_{i,j}-v_{i+1,j})+\rho_{s}(v_{i,j} - v_{i,j+1})]
\end{dmath}
where u and v are respectively horizontal and vertical component of velocity for frame $I_{1}$ and $I_{2}$,$\lambda$ is a regularization parameter based on expected smoothness of the flow field and 
$\rho_{D}$, $\rho_{s}$ are two functions. 
Values of u and v are calculated by optimizing E(u,v) term. 

\begin{figure}[h!]
  \centering
      \includegraphics[width=0.5\textwidth,height=3.5in]{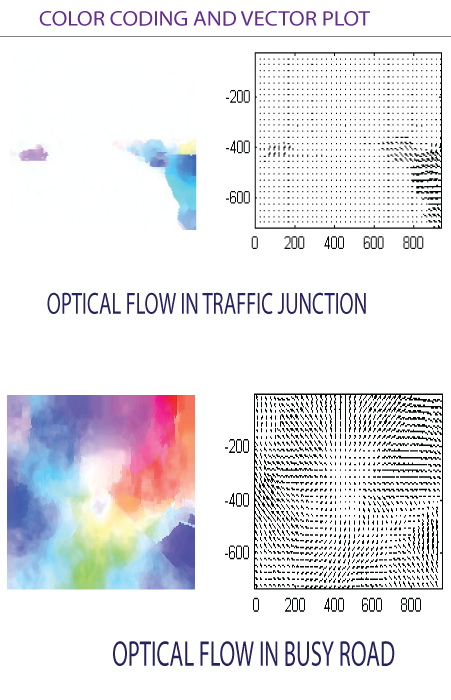}
  \caption{Optical flow characteristics}
\end{figure}

\subsection{Wrong direction driving}
In Indian conditions, vehicles coming in wrong direction is also another 
frequent case of rash driving or rather nuisance. Wrong direction driving 
is easily estimated by observing anomolies in optical flow in lanes. 

\subsection{Frequent lane change detection}
Another characteristic of rash drivers is frequent lane changing. 
We have used a robust illuminant invariant lane detection system in our
work using inverse perspective projection \cite{Bertozzi1996} and
cubic interpolation with RANSAC curve fitting \cite{Fischler1981}.
We have assumed a parabolic road model. From road lane fitting, relative 
position of other vehicles with respect to lanes can be estimated. 
Also from our lane detection algorithm, deperture angle of test car 
from current lane can be computed.

\begin{figure}[h!]
  \centering
      \includegraphics[width=0.5\textwidth]{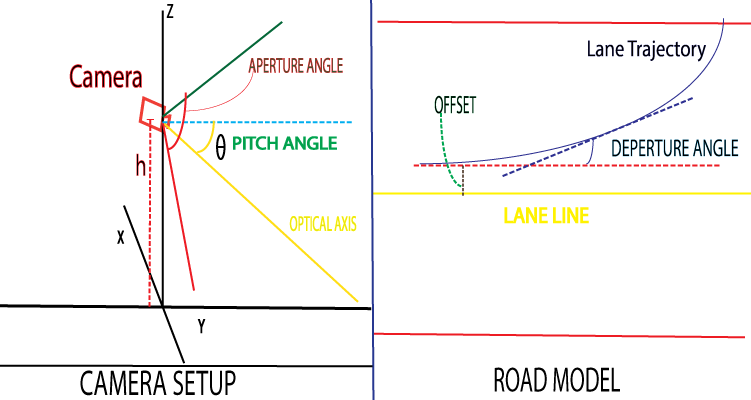}
  \caption{Camera setup and Road model}
\end{figure}
We have used Lab color space for seperating color and illuminant part of images for better detection of lane lines using 2nd adn 4th order steerable filters. In Fig.[4] we presented camera setup and assumed road model. 
 \begin{figure}[h!]
  \centering
      \includegraphics[width=0.5\textwidth,height=2in]{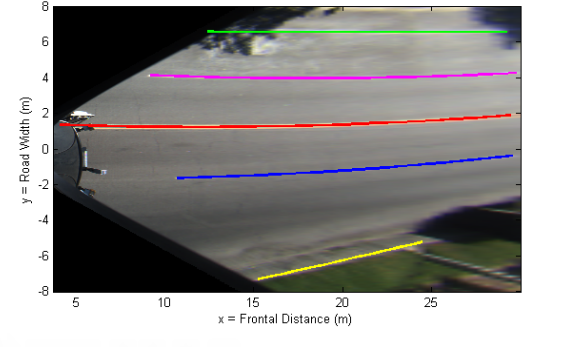}
  \caption{Lane detection algorithm result}
\end{figure}

\subsection{Driving-close-to-vehicles and people-in-front behaviour}
Not maintaining a proper distance from nearby cars or people in pedestrains is also a facet of rash-driving.  
We have used HOG feature based deformable part model for detecting and locating other cars and pedestrians 
with respect to lane lines. For detecting and locating object in image we wil use pyramid based template 
matching method, where we train car and person model using deformable part model \cite{fel2009,fel2008} 
based on HOG \cite{dalal2005} feature. This method can detect car and people very efficiently under occlusion 
also. Latent SVM trained model is shown in Fig.[6]; and detected car and people is shown 
in Fig.[7] (Reference for the images \cite{dataset2013}). 
\begin{figure}[h!]
  \centering
      \includegraphics[width=0.5\textwidth]{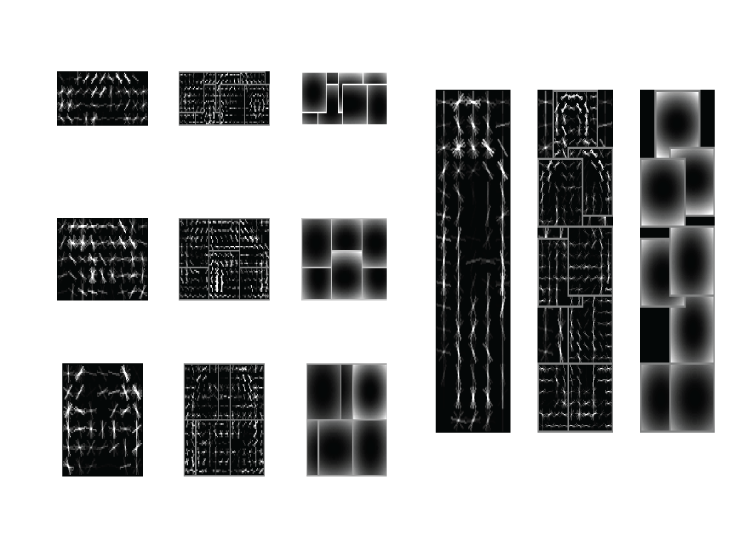}
  \caption{Root car and person models and its part,\cite{fel2009,fel2008}}
\end{figure}
\begin{figure}[h!]
  \centering
      \includegraphics[width=0.5\textwidth]{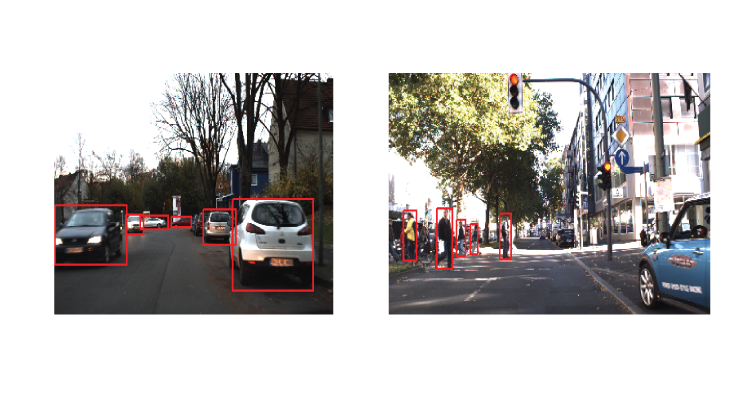}
  \caption{Car and person detected using above deforable model}
\end{figure}


\subsubsection{Pinhole Camera Model}
For determining distance of obstacle from test car we will use pinhole camera model, this model can give good accuracy for object in front of test car. From error analysis we set different offset for approximately measuring distance.
 
If a 3D point  P= $ (u,v,w)^{T} $ and its pinhole projected point is $ I_{p} =  (x,y)^{T} $ , there relation\cite{princeCVMLI2012} is given by following equation
\begin{dmath}
x = \frac{\phi_{x}(\omega_{11}u+\omega_{12}v+\omega_{13}w+\tau_{x}) + \lambda(\omega_{21}u+\omega_{22}v+\omega_{23}w+\tau_{y})}{\omega_{31}u+\omega_{32}v+\omega_{33}w+\tau_{z}} + \delta_{x}
\end{dmath}
\begin{dmath}
y = \frac{\phi_{y}(\omega_{21}u+\omega_{22}v+\omega_{23}w+\tau_{y})}{\omega_{31}u+\omega_{32}v+\omega_{33}w+\tau_{z}} + \delta_{y}
\end{dmath}
where intrinsic matrix $\Lambda$ is given by
\[
\Lambda = 
	\begin{bmatrix}
		\phi_{x} & \lambda & \delta_{x}\\
		0 & \phi_{y} & \delta_{y}\\
		0 & 0 & 1\\
	\end{bmatrix} 
\]

Rotation matirix of camera can be given by
\[
\Omega = 
	\begin{bmatrix}
		\omega_{11} & \omega_{12} & \omega_{13}\\
		\omega_{21} & \omega_{22} & \omega_{23}\\
		\omega_{31} & \omega_{32} & \omega_{33}\\
	\end{bmatrix} 
\]

Translation matix as 
\[
\tau = 
	\begin{bmatrix}
		\tau_{x}\\
		\tau_{y}\\
		\tau_{z}\\
	\end{bmatrix} 
\]


Finally, employing all the features described, we make a
estimate of rash-driving. For now, our proposed system is
rule-based. In the future, we will collect samples with and
without rash driving, using professional drivers. Using a
generative machine learning approach, we can build a probabilistic model to predict rash-driving. We are also considering recording data with naive volunteers, and manually annotating parts of the data where rash-driving tendencies
are seen, so as to validate the model.

\section{Conclusions}
In this paper we have described a visual analyis method to characterize driving 
behavior, with a specic focus on rash-driving. Our algorithm is based on 
calibrated single camera images. The methods are general enough to work 
on cameras placed on cars as well as on infrastructure. In the future, 
we plan to integrate this module with the automatic number plate detection and recognition
module (as in \cite{christos2006}) for a traffic monitoring application. 
Though our work is ongoing and preliminary, we believe such a system can have 
a good societal impact. As described in Section 1 Introduction, we will record 
a rash-driving dataset in Indian conditions, and test our methods. 
We will also make a requirements study with government agencies. 

\bibliographystyle{abbrv}
\bibliography{submis}  

%
%

\end{document}